\documentclass[10pt,twocolumn,letterpaper]{article}

\usepackage{cvpr}
\usepackage{times}
\usepackage{epsfig}
\usepackage{graphicx}
\usepackage{amsmath}
\usepackage{amssymb}
\usepackage{color}

\usepackage[breaklinks=true,bookmarks=false]{hyperref}

\cvprfinalcopy 

\def\cvprPaperID{****} 

\begin{document}

\title{Improvements to context based self-supervised learning}

\author{T. Nathan Mundhenk\\
Computational Engineering Division\\
Lawrence Livermore National Laboratory\\
Livermore, California\\
{\tt\small mundhenk1@llnl.gov}
\and
Daniel Ho\\
EECS Department\\
University of California, Berkeley\\
Berkeley, California\\
{\tt\small daniel.ho@berkeley.edu}
\and
Barry Y. Chen\\
Computational Engineering Division\\
Lawrence Livermore National Laboratory\\
Livermore, California\\
{\tt\small chen52@llnl.gov}
}

\maketitle

\widowpenalty10000
\clubpenalty10000
\definecolor{myred}{RGB}{255, 0, 0}
\definecolor{myblue}{RGB}{0, 0, 255}

\begin{abstract}We develop a set of methods to improve on the results of self-supervised learning using context. We start with a baseline of patch based arrangement context learning and go from there. Our methods address some overt problems such as chromatic aberration as well as other potential problems such as spatial skew and mid-level feature neglect. We prevent problems with testing generalization on common self-supervised benchmark tests by using different datasets during our development. The results of our methods combined yield top scores on all standard self-supervised benchmarks, including classification and detection on PASCAL VOC 2007, segmentation on PASCAL VOC 2012, and ``linear tests'' on the ImageNet and CSAIL Places datasets. We obtain an improvement over our baseline method of between 4.0 to 7.1 percentage points on transfer learning classification tests. We also show results on different standard network architectures to demonstrate generalization as well as portability. All data, models and programs are available at: \url{https://gdo-datasci.llnl.gov/selfsupervised/}.     
\end{abstract}

\section{Introduction}

Self-supervised learning has opened an intriguing new avenue into unsupervised learning. It is intellectually satisfying due to the way it resembles gestalt-like mechanisms of learning in visual cortical formation. It is also appealing for that fact that it can be implemented with standard off-the-shelf neural networks and toolkits.

Self-supervised learning methods create a protocol whereby the computer can learn to teach itself a supervised task. For instance, in \cite{Doersch15} a convolutional neural network (CNN) was taught to learn the arrangement of patches in an image. By learning the relative position of these patches, the network would be forced to also learn the features and semantics that underlie the image. Although the network is trained to learn patch positions, the final goal was to generalize the learned representation to solve other tasks. For instance, the self-supervised network was trained on a transfer task (fine-tuned) to classify objects in the PASCAL VOC dataset \cite{Everingham10}, and compared with a CNN trained on a supervised task, such as learning to classify the ImageNet dataset \cite{Imagenet09}. If the self-supervised network learned good generalizations of image features and semantics, it should perform as well as a supervised network on transfer learning. 

Over the last few years, several methods of self-supervised learning have been introduced. For instance, \cite{Dosovitskiy15} trained a CNN to recognize which transformation had been performed on an image. Since then, methods have been introduced that use context arrangement of image patches \cite{Doersch15,Noroozi16a}, image completion \cite{Pathak16}, image colorization \cite{Zhang17,Zhang16,Larsson17}, motion segmentation \cite{Pathak17}, motion frame ordering \cite{Wang15b,Wang17,Lee17}, object counting \cite{Noroozi17} and a multi-task ensemble of many models \cite{Doersch17,Kim18}. 

Each method has relative strengths and weaknesses. For instance, \cite{Zhang17} uses a customized ``split-brain'' architecture that makes it less off-the-shelf than other solutions which frequently use a standard network such as AlexNet \cite{AlexNet}. \cite{Wang15b,Wang17,Lee17,Pathak17} all use motion cues, but have the downside of being constrained to video data.  Many patch based methods use a Siamese network architecture which incurs extra memory demands \cite{Doersch15,Noroozi16a,Noroozi17,Lee17}. However, every self-supervised method suffers the drawback of still being very short of supervised methods in terms of transfer learning performance. 

Our intent in this work is to improve the performance of self-supervised learning. We present a variety of techniques we hope are applicable to other approaches as well. We will demonstrate generalizability of these techniques by running them on several different neural network architectures and many kinds of datasets and tasks.  For instance, as we will discuss, one dataset we wish to add to the corpus of standard self-supervised tests, the Caltech-UCSD Birds 200 set (CUB) \cite{Welinder10} is excellent for finding potential short comings of techniques designed to address the well-known \emph{chromatic aberration} problem \cite{Doersch15}. We will suggest this is due to the importance of color patterns in bird classification \cite{Cornell09}.  


\section{Issues and Related Work}
\begin{figure*}
\centering
\includegraphics[height=3.75 cm]{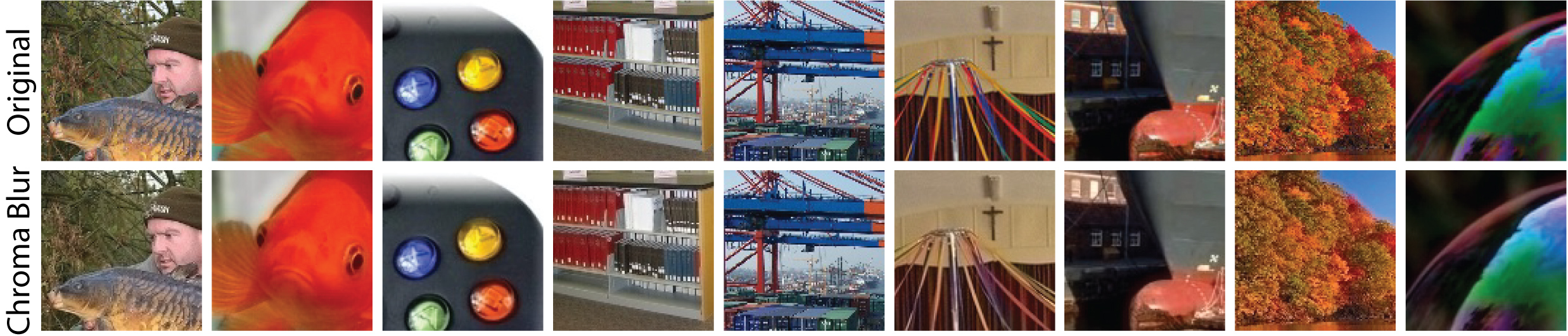}
\caption{These are examples of patches taken from ImageNet that are used during self-supervised training. Below each original is an example with chroma blurring. It is frequently difficult to distinguish the blurred from original images, because humans are not very spatially sensitive to variation of color. Chroma blurring can sometimes result in a loss of color saturation and color bleeding of very saturated regions (such as the red ship bow, third from right). Notice the original gold fish image has signs of strong chromatic aberration (top-left of head). This is blended out effectively by chroma blurring which switches the green aberration to the fish's own red color. \emph{See supplementary material appendix} figure ~\ref{fig:chroma_blur_layer} for conv1 layer filter comparisons. }
\label{fig:chroma_blur}
\end{figure*}
We use a patch/context approach for the issue of self-supervised learning \cite{Doersch15,Noroozi16a}. This is a popular method, but is by no means the only active path of inquiry. Patch/context approaches work by creating an arrangement of image patches in either space or time. Each distinct arrangement is assigned a class label, and the network then predicts the correct arrangement of these patches by solving a supervised classification problem. The network can be a typical supervised network such as AlexNet \cite{AlexNet}, VGG \cite{VGGNet}, GoogLeNet \cite{GoogLeNet} or ResNet \cite{ResNet}. In order to view multiple patches at the same time, a Siamese network is frequently used where each patch is fed into an independent network path, and each path shares weights. \cite{Doersch15} used a system of two patches in a finite set of eight possible spatial configurations. \cite{Noroozi16a} created an extension using as many as nine patches in a \emph{puzzle} configuration. Temporal ordering of patches can also be used. For instance, \cite{Lee17} shuffled four consecutive video frames to create 12 classes for prediction. Another temporal method \cite{Wang17} queried the networks ability to determine if a patch came from the same object later in time or a similar but different object. This can be considered a meta self-learner since it leverages \cite{Doersch15} as a pre-self-supervised learner to determine object similarity.

Patch based methods have the advantage of being easy to understand, network architecture agnostic, and frequently straightforward to implement. They also tend to perform well on standard measures of transfer learning. For instance, \cite{Doersch15} is a top performer on PASCAL VOC 2007 detection \cite{Girshick15}, even among a large number of new arrivals. \cite{Noroozi16b} is almost tied for the top score on PASCAL VOC 2007 classification \cite{Krahenbuhl16} and has the top score for PASCAL VOC 2007 detection and the second highest score for PASCAL VOC 2012 segmentation \cite{Long15,Pathak16}.

However, patch/context-based networks typically suffer from an issue of being able to ``cheat'' and bypass learning the desired semantic structure by finding incidental clues that reveal the location of a patch. An example of this is \emph{chromatic aberration}, which occurs naturally as a result of camera lensing where different frequencies of light exit the lens at slightly different angles. This radially offsets colors such as magenta and green modestly from the center of the image. By looking at the offset relation of different color channels, a network can determine the angular location of a patch in an image. Aberration varies between lenses, so it's an imperfect cue, but one that none-the-less exists. A common remedy is to withhold color data from one or more channels by using channel-dropping \cite{Doersch15,Doersch17}, channel replication \cite{Lee17}, or conversion to gray scale \cite{Noroozi17}.  The primary difficulty with these approaches is that color becomes decorrelated (or absent) since colors are not observed together. This makes it difficult to learn color opponents for patterns that emerge in supervised training sets, such as ImageNet. Another approach is to jitter color channels \cite{Noroozi16a}, but this has a similar effect to blurring an image, and it might affect the sharpness of learned wavelet features.

An often-cited worry in all patch/context works relates to trivial low-level boundary pattern completion \cite{Doersch15,Noroozi16a,Doersch17,Noroozi17,Lee17}. The neural network may learn the alignment of patches not based on (for instance) desirable semantic information, but instead by matching the top or bottom part of simple line segments. Two common approaches are to provide a large enough gap between patches and to randomly jitter the patches. This last technique may be dubious since a convolutional neural network can align simple patterns at arbitrary offsets. This issue may also be implicitly addressed by having non-4-connected adjacent patches. Half the patches in \cite{Doersch15} are arranged diagonally which should make them resistant to trivial low-level boundary pattern completion. Also, we should note that while we would not want a self-supervised learner to use this cheat all the time, it could be used as a cue to help form low level features. So, it is somewhat unclear how much of a problem this might be. 

In another \emph{possible} problem for self-supervised networks in general, mid-layers in the network may not train as well as the early and later layers.  For instance, \cite{Doersch17} created a self-supervised network using an ensemble of different methods. They then created an automated lasso to grab layers in the network most useful for their task. The lasso tended to grab layers very early or very late in the network. This suggests that for many self-supervised tasks, the information in the middle network layers is not very essential for training. Another piece of evidence comes from the CSAIL Places linear test \cite{Zhang17,Zhang16}, which shows how well each layer in the network performs on transfer learning. Many self-supervised networks perform as well or better than a supervised ImageNet trained network at the first and second convolutional layers in AlexNet, but struggle at deeper layers. 


\section{Approach}

Our approach is comprised of three parts. The first is a collection of tools and enhancements which for the most part should be transferable to other self-supervised methods. In our second part, we utilize two new datasets to make our experiments more diverse and general. The third part is a demonstration on several different neural networks to verify generalization and demonstrate portability.

For our general approach, we start with the baseline of \cite{Doersch15} using a two-patch spatial configuration paradigm. This approach gives good results, and is easy to both implement and understand. We then augment this approach using various techniques. For each technique, we vigorously test effects empirically to justify their usage.

\subsection{Our Toolbox of Methods}

\subsubsection{Chroma Blurring (CB)}

We address the problem of chromatic aberration by removing cues about image aberration while allowing color patterns and opponents to be at least partially preserved. We note that the human visual system is not very sensitive to spatial frequency of chroma, but is much better at discerning detail about shifts in intensity \cite{Livingstone02}. However, even with this lack of spatial acuity for color, we can still discern meaningful color patterns. As such, we balance the tradeoff between decreasing color spatial information and removing chromatic aberration cues.

To preserve intensity information, but reduce aberration, we start by converting images into \emph{Lab} color space. We then apply a blurring operation to the chroma \emph{a} and \emph{b} channels. In this case, we use a 13x13 box filter. It is two times the size of the 7x7 convolution filter of our original GoogLeNet target network. The luminance channel is left untouched, and we convert the image back to RGB. Figure~\ref{fig:chroma_blur} shows several patches which we have \emph{chroma blurred} for comparison.

\subsubsection{Yoked Jitter of Patches (YJ)}

Most patch/context methods apply a random jitter between patches \cite{Doersch15,Noroozi16a,Doersch17,Lee17}. The different patches are jittered in different amounts and different directions.  One issue with applying a random jitter is that it might distort or skew the spatial understanding in the network. As an example, if the head of an animal is observed in one patch and the feet in the other, the true spatial extent between these items would be difficult to discern given a random jitter: the patches might be 30 pixels apart, or they might be 60. If on the other hand, the patches maintained a fixed spacing, reasoning about the extent of an object between patches would be easier. Thus, the network might make better inferences about the larger shape of an object beyond each patch itself. 

We do this by yoking the patch jitter. Each patch is randomly jittered to create a random crop effect, but they are jittered by the same amount in the same direction. This might make us prone to trivial low-level boundary pattern completion, but as mentioned, we suspect this can be partially addressed by having non-4-connect patches. Also, it is unclear how well a random jitter will address such a problem since features do not need to be aligned in order to be recognized in a CNN. Additionally, a certain amount of low-level boundary pattern completion may not be a problem since it may enhance learning of simple features. 
\begin{figure*}
\centering
\includegraphics[height=8.4 cm]{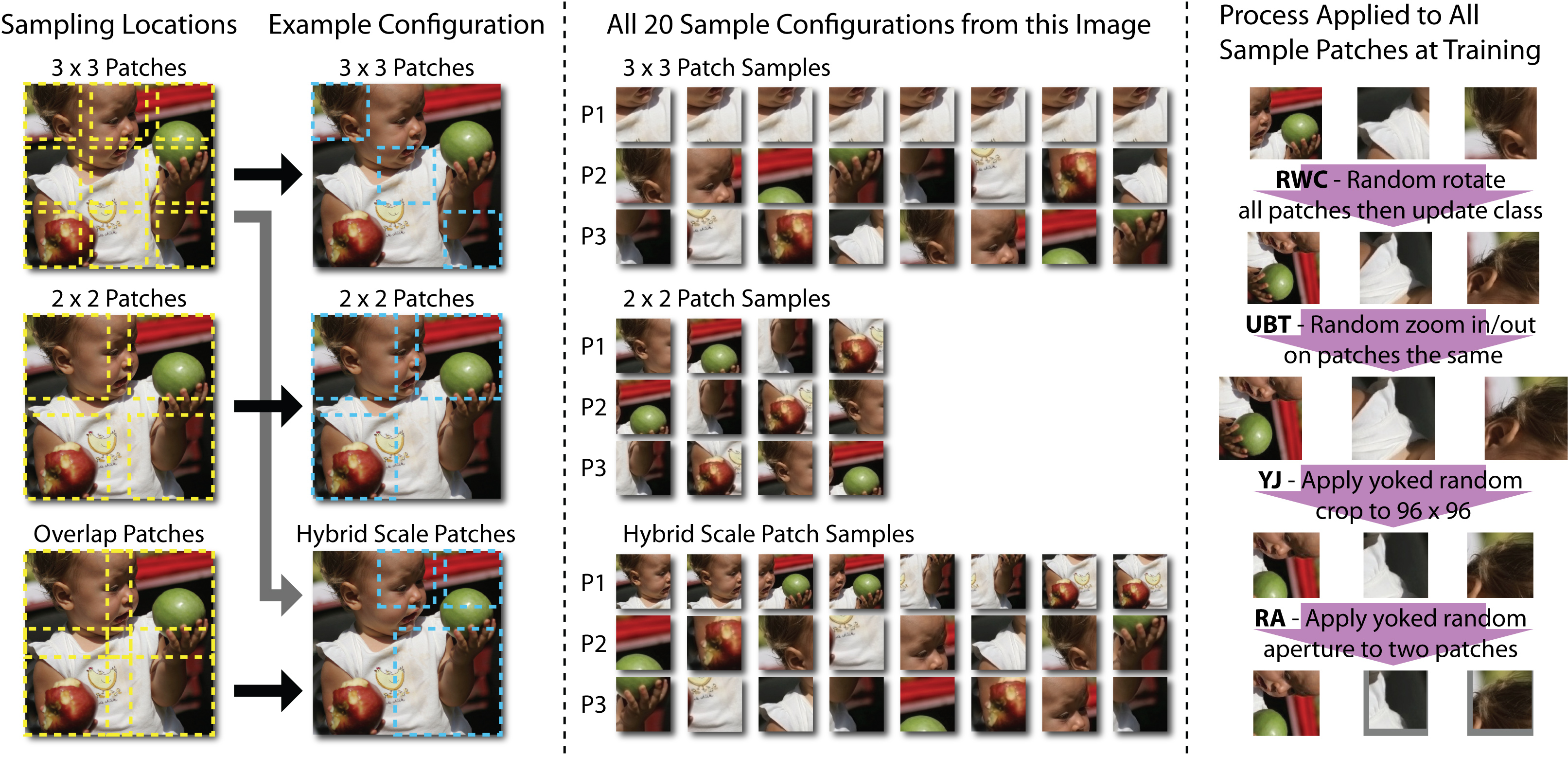}
\caption{The left column shows the location patches are extracted from in the image. The next column over shows some example configurations obtained from those patches. In the middle are all 20 patches extracted from this (and every) image. The order is labeled for each patch in a set as \emph{P1},\emph{P2} and \emph{P3}. The right column shows how these patches are fed into the process to then create the final patches fed into the neural network.}
\label{fig:patch_example}
\end{figure*}
\subsubsection{Additional Patches and Configurations}

We use the same 96x96 sized patches as \cite{Doersch15}, since it fits the receptive field of a 3x3 convolution at the end of many CNNs. That is, most of the popular networks have a five-layer topology of layer groups with each layer group being half the dimension of the preceding one \cite{GoogLeNet,ResNet,DenseNet,AlexNet,Mundhenk2016} (AlexNet technically has four group layers, but the first layer has half-scale of most other networks). There is some dimensional variation caused by the omission of padding in some layers, but one pixel on the end layer maps to an extent of about 32 pixels in the input image. Since most networks tend to only use 3x3 convolutions at the last layers, a patch size of 96x96 is justified to cover its full field.

We use three patches {\bf (TP)} in each set. In the two-patch configuration of \cite{Doersch15}, one of the two patches may not cover an area with useful information. For instance, half of the image may be covered by ocean. We can address this problem by using more patches. \cite{Noroozi16a} uses a ``puzzle'' system of nine 80x80 patches. However, if we want to use 96x96 sized patches and be able to train larger, more contemporary networks, a 9x9 puzzle may not be feasible to train. If, on the other hand, we just add one more patch, we create a triple Siamese network which is smaller than a single network over the traditional 224x224 image size. This makes it easy to move to a larger network. We did not test four patches.

We add extra patches configurations {\bf (EPC)}. That is, we added several new configurations of patches seen in Figure~\ref{fig:patch_example}. Adding new patterns (1) creates more orthogonal and unique patterns (2) covers more of the image at once (3) mixes scales to prevent simple pattern completion (4) creates a natural way to cover the image, but use multiple scales for training.

We draw three different patterns of patches. We start by extracting all patches at a 110x110 resolution. \emph{3x3} patches are taken from a 384x384 image. This is taken from the center of an aspect preserved image by reducing the smallest dimension to that size. The patches are evenly spaced and aligned with the image corners to cover the image (there is no edge margin). \emph{2x2} patches are taken from a 256x256 image and \emph{overlap} patches are taken from a 196x196 image. We then use eight 3x3 patterns similar to \cite{Doersch15}, 2x2 patterns are L-shaped and we use four of them. We combine patches from 3x3 and overlap patches to create \emph{hybrid} patch sets. These are hybrid scale patches which allow more semantic reasoning by preventing easy matching of simple features between patches since they are at different scales. The processed patches are fed into the network in batches. The final patches fed into the network are sized 96x96. Note that Figure~\ref{fig:patch_example} shows all 20 typical combinations from a sample image. These are the exact same patterns extracted from all images in the ImageNet training set. In all, we obtain 25,623,340 training patch sets from ImageNet. 

\subsubsection{Random Aperture of Patches (RA)}
\begin{figure}
\centering
\includegraphics[height=3.0 cm]{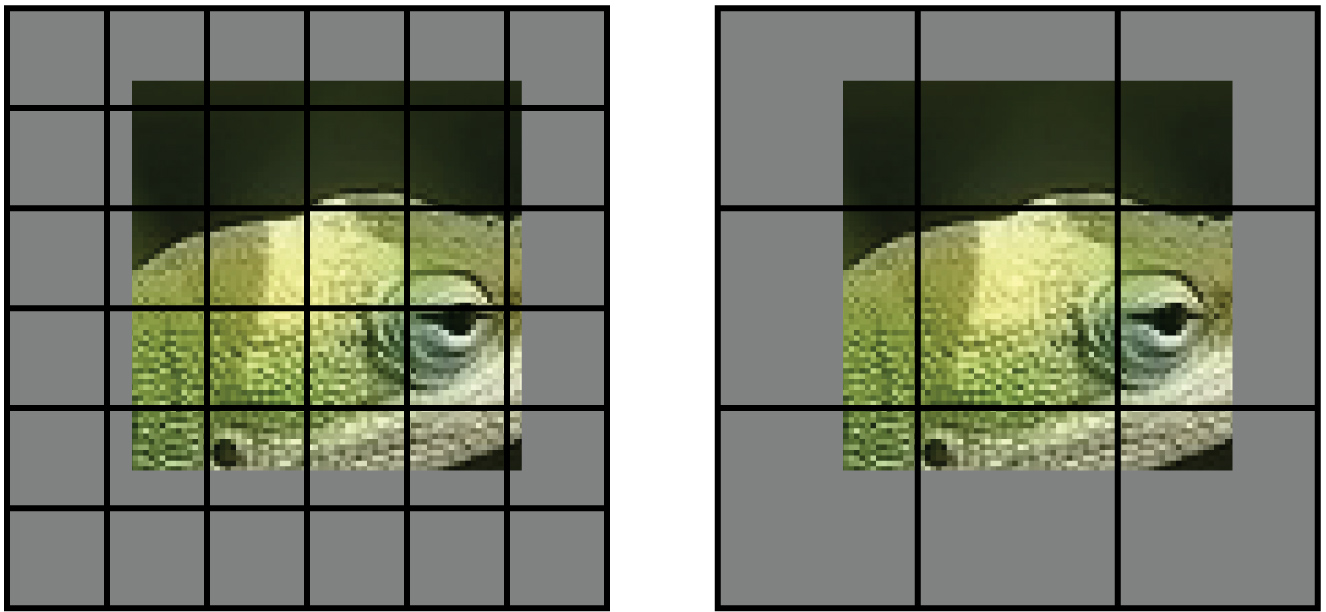}
\caption{The grayed-out area has been apertured on a 96x96 size patch. The aperture is 64x64, the smallest size we use. The left image shows the pixel arrangement on group layer four. At least one 3x3 region is not directly interfered with by the aperture. However, on the right, layer 5, only one pixel is fully uncovered. All spatial interactions at this layer will involve at least one occluded region. Ideally, this would create some inhibition to layer five forming meaningful spatial associations and perhaps bias towards layer 4 which can. Note that this description is simplified and somewhat imprecise since image information can propagate laterally through consecutive layers.}
\label{fig:aperture}
\end{figure}
We mentioned some (minor) evidence that middle layers in a self-supervised network are being neglected. One approach would be to try and create a bias towards these neurons. In the general five-layer topology, the fourth group layer has a receptive field of 48x48 given a 3x3 filter. In AlexNet, this would include layers conv3, conv4 and conv5. In GoogLeNet, this would include all 4th layers (4a, 4b etc). If we create an aperture, we could create a patch that doesn't cover the extent of the 5th group layers, but does cover the extent of the 4th group layers filters. This could bias against the 5th layers from learning since it cannot see the whole patch. Ideally, this would put emphasis on learning in the 4th layers. See Figure~\ref{fig:aperture} for an example of this.

A random aperture on two of the three patches in a set is created. The idea behind leaving one patch un-apertured is so that we don't completely bias against group layer 5, we still want it to learn. The aperture is square and for each sample is randomly sized between 64x64 and 96x96. The minimum size is 64 since this is the smallest size we can use and guarantee that at least one 3x3 convolution is unobstructed in the fourth layer. The position of the aperture is also randomized but must fit inside the patch so we can never have a viewable area less than 64x64.  The area outside the aperture is filled with ImageNet mean RGB. The size and position of the aperture in two patches is yoked. Which two patches are apertured is randomized for each sample. 

\subsubsection{Rotation with Classification (RWC)}

\begin{figure}
\centering
\includegraphics[height=3.1 cm]{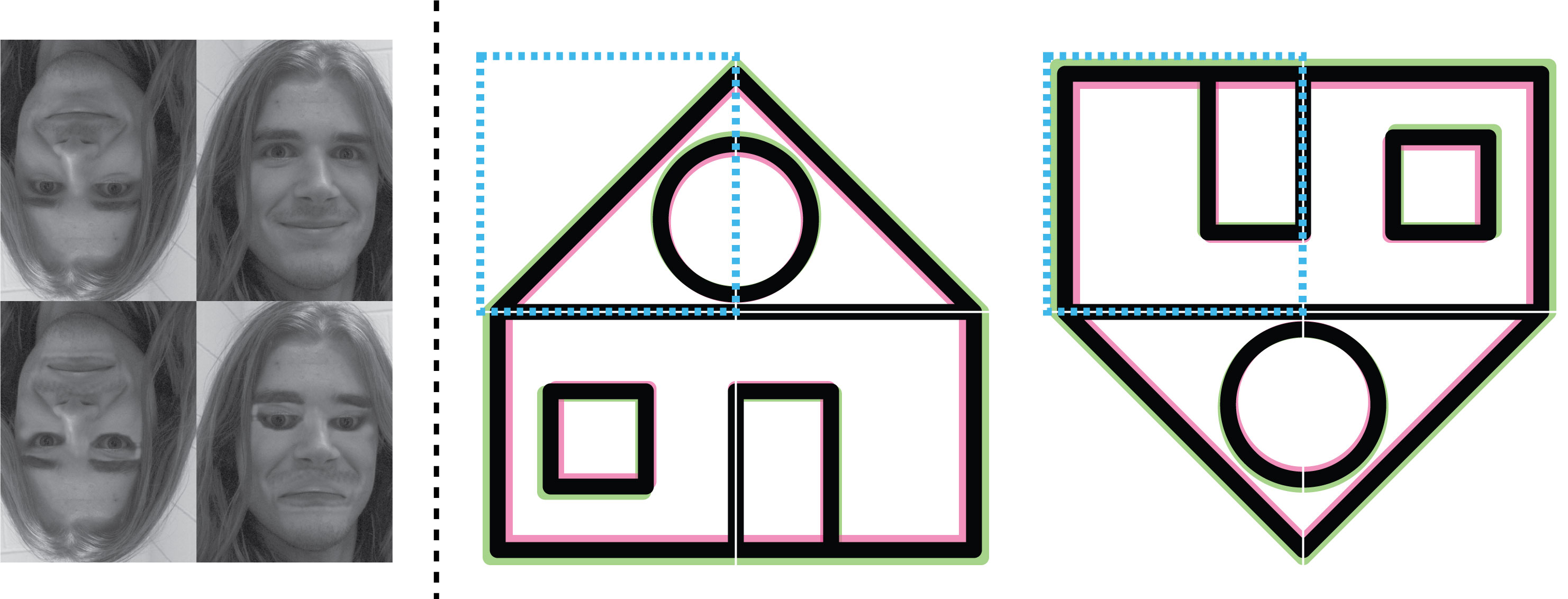}
\caption{On the left is an example of the famous thatcher illusion \cite{Thompson80,MTeffect06}. It demonstrates conditional sensitivity to upside-down features in an image against the background. We used this mostly as inspiration. On the left house image \cite{ChromaHouse}, the network can tell that the blue bordered area comes from the upper left corner based on chromatic aberration alone. However, on the right image, rotation with classification makes it tell us if the patch is inverted and comes from the lower right corner. If it uses chromatic aberration as the only cue, it would be wrong 50\% of the time. (Figure is enlarged in appendix: see figure ~\ref{fig:rwc_big})}
\label{fig:rwc}
\end{figure}
Each patch in a patch/context model may simply contain a part of a much larger object. In general, this is the intent of the patch/context approach. Might it help if parts can be understood at different orientations? For instance, if one has seen an upside-down roof top, one may better understand a triangular yield sign or a funnel. Additionally, humans have the ability to \emph{conditionally} recognize upside-down parts embedded in a whole image. This is illustrated by the famous Thatcher illusion \cite{Thompson80} (see Figure~\ref{fig:rwc}). We reason that self-supervised learning might benefit from exposure to upside-down patches, and it would help to make the network identify if patches are right-side-up or upside-down. We do this by flipping the whole image so that all patches are flipped. Then, we double the number of classes by giving each upside-down image its own class. For instance, if we have 20 classes of patch arrangements, when we add upside-down images, we have 40 classes. We also explore 90 and 270 degree rotations. This yields a total of 80 classes.

Forcing the network to classify patches as upside-down also reduces the strength of clues generated by chromatic aberration. Aberration radiates from the center of the image. Without rotating the image, a downward sloping arch of green/magenta to the left indicates the patch comes from the upper left-hand corner. However, in a flipped image, the same pattern indicates the lower right corner instead. By just trying to guess upper left-hand corner from the chromatic aberration pattern, it will be wrong 50\% of the time. With four rotations, it will be wrong 75\% of the time.   

\subsubsection{Miscellany}
  
 We present experiments with a few other tricks which we found helpful to varying degrees. One method is a typical mixture of label preserving transformations \cite{Simard03,Ciresan11,Ciresan12} we are calling the \emph{usual bag of tricks} {\bf (UBT)}. This involves augmentation by randomly mirroring, zooming, and cropping images. The mirroring is simple horizontal flipping and has no special classification, like with RWC, since this would most likely prove confusing to the network. For random zooming, we randomly scaled each input 110x110 patch to between 96x96 to 128x128, and then extract a random 96x96 patch from this. The zoom and crop location is random for each sample, but is yoked between the three input patches in a set.

Borrowed from \cite{Noroozi17}, we take the idea of mixing the method of rescaling during UBT. Each of the three patches in a set is rescaled by one of four randomly chosen rescale techniques (Bilinear, Area, Bicubic or Lanczos). The random selection is \emph{not} yoked between patches. The idea is yet again to make it harder to match low level statistics between patches (trivial solutions). We call this randomization of rescaling methods {\bf (RRM)}. 

We also tried varying the learning rate and decay rate of network layers to increase learning of middle layer weights. For instance, one can adjust the first layer to have 70\% the learning rate as the center most layer. We try to linearly increase the learning rate towards a middle layer and then reduce the rate back down. Given the nine layers of a Siamese AlexNet, we would have learning rates \{0.7, 0.8, 0.9, 1.0, 0.9, 0.8, 0.7, 0.6, 0.5\}. Here, conv4 layer has learning rate multiplier 1.0 and the last fully connected layer has 0.5. We call this weight varying {\bf (WV)}. 

\subsection{Verification Datasets}

The development and testing of new techniques generally requires fishing for results. As such, one should avoid using the target dataset for testing each new idea. Fishing leads to solutions specialized towards the specific dataset rather than a general solution. For self-supervised learning, test metrics based on PASCAL VOC \cite{Everingham10}, ImageNet \cite{Imagenet09} and CSAIL Places \cite{Places14} are commonly used. Therefore, we test techniques on a few new datasets with a certain amount of overlap, but which possess differences so that we can be more confident in generalization. 

For validation, we use a combination of CUB birds \cite{Welinder10} (a fine-grained bird species dataset) and CompCars \cite{Yang15} (a fine-grained car model dataset). We call this combination CUB/CCars (Examples from these sets are in the appendix as figures~\ref{fig:birds} and ~\ref{fig:compcars}). We use these sets by training a network in a self-supervised manner and then apply transfer learning by fine-tuning them for classification. Both data sets are fine grained for their respective class (birds and cars). However, there are major differences between the kinds of features cars and birds have. Additionally, the CUB birds dataset provides an ideal test set for dealing with chromatic aberration. The four keys to identifying birds are size/shape, habitat, behavior and color pattern \cite{Cornell09}. When trying to control chromatic aberration, one may alter the image in a way that negatively affects color processing and thus classification for birds.

\subsection{Alternative Networks}

\begin{figure}
\centering
\includegraphics[height=5.6 cm]{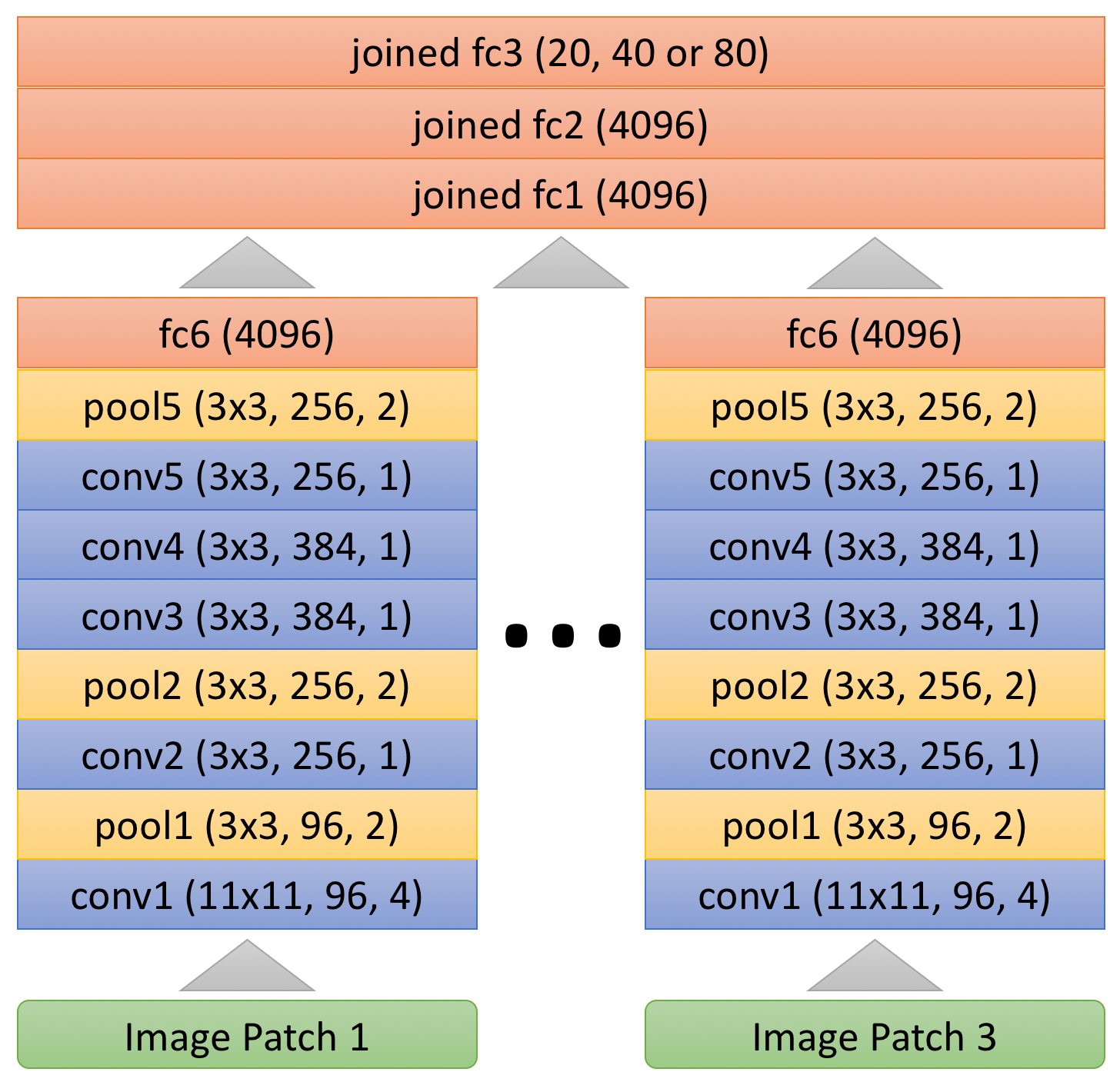}
\caption{This is our custom batch normalized triple Siamese AlexNet. It is very similar to \cite{Doersch15}. Each layer has a batch norm layer after it. Notice we have removed LRN \cite{AlexNet} layers.}
\label{fig:alexnet}
\end{figure}
We are interested in generalization of our solutions and portability to other architectures. If we constrain self-supervised learning to mostly being a training protocol, it's easier to train on different networks. As with using many datasets, using many networks also helps to assure that a technique is not network specific, but works well on other designs. We demonstrate results on four different networks. These are (a) standard CaffeNet type AlexNet \cite{AlexNet,Caffe} (b) AlexNet with \emph{batch normalization} {\bf (BN)} \cite{BatchNorm} (figure ~\ref{fig:alexnet}) (c) a ResCeption network \cite{Mundhenk2016} (d) an Inception network with BN \cite{Inceptionv4,BatchNorm}.

\begin{table*}
\begin{center}
\footnotesize
\begin{tabular}{l|lllll|lllll}
\hline\noalign{\smallskip}
\multicolumn{1}{c}{} & \multicolumn{5}{c}{{\bf Accuracy}} & \multicolumn{5}{c}{{\bf Improvement}} \\ 
\hline\noalign{\smallskip}
{\bf Method} &	{\bf CUB} &	{\bf CCars} &	{\bf Mean} &	{\bf VOC} &	{\bf All} &	{\bf CUB} &	{\bf CCars} &	{\bf Mean} &	{\bf VOC} &	{\bf All}\\
\noalign{\smallskip}
\hline
\noalign{\smallskip}
No Pretrain & 	56.20 &	59.13 &	57.67 &	55.12 &	56.82	& -- & -- & -- & -- & --\\				
ImageNet Supervised &	74.44 &	85.40 &	79.92 &	72.67\textcolor{myblue}{\textsuperscript{\S}} &	77.50 & -- & -- & -- & -- & --\\\noalign{\smallskip}
\hline
\noalign{\smallskip}					
Baseline 2 Patch Protocol &	62.33 &	79.86 &	71.09 &	63.61 &	68.60	& -- & -- & -- & -- & --\\			\noalign{\smallskip}
\hline
\noalign{\smallskip}	
add Chroma Blurring {\bf (CB)} &	64.29 &	80.80 &	72.55 &	64.98 &	70.02 &	1.97 &	0.94 &	1.45 &	1.37 &	1.42\\
add Yoked Jitter {\bf (YJ)} &	65.17 &	80.95 &	73.06 &	65.15 &	70.42 &	0.87 &	0.15 &	0.51 &	0.16 &	0.40\\
add 3rd Patch {\bf (TP)} &	65.19 &	81.54 &	73.36 &	65.27 &	70.66 &	0.02 &	0.59 &	0.30 &	0.12 &	0.24\\ 
add Extra Patch Cfgs. {\bf (EPC)} &	67.07 &	80.50 &	73.79 &	65.67 &	71.08 &	1.89 &	\textcolor{myred}{-1.04} &	0.43 &	0.41 &	0.42\\
add Usual Tricks {\bf (UBT)} &	67.91 &	80.83 &	74.37 &	65.58 &	71.44 &	0.84 &	0.33 &	0.58 &	\textcolor{myred}{-0.10} &	0.35\\
add Rand. Aperture {\bf (RA)} &	68.01 &	82.07 &	75.04 &	66.79 &	72.29 &	0.10 &	1.24 &	0.67 &	1.21 &	0.85\\
add Rotation 180 {\bf (RWC)} &	68.89 &	84.23 &	76.56 &	68.39 &	73.83 &	0.88 &	2.16 &	1.52 &	1.60 &	1.55\\
add Rotation 90, 270 and WV &	69.39 &	84.25 &	76.82 &	68.31 &	73.99 &	0.50 &	0.02 &	0.26 &	\textcolor{myred}{-0.07} &	0.15\\
\hline
\end{tabular}
\end{center}
\caption{
This is a basic ablation showing the effect of adding each method one at a time. The scores for CUB birds (CUB) and CompCars (CCars) are the single class classification accuracy. PASCAL VOC uses mean average precision (mAP). The mean column is for CUB and CCars, but we show a mean of CUB, CCars and VOC as ``All''. VOC is the \emph{Post Hoc} classification results run after the fact to see how well our CUB/CCars surrogate set matches a core self-supervised benchmark test. The baseline two patch protocol uses color dropping and matches the protocol of \cite{Doersch15}. Gains in CUB/CCars appear to correlate with gains in VOC (but not perfectly). The largest gains for both CUB/CCars and VOC are from rotation with classification, chroma blurring and random aperture.  Also notice that the results for CompCars is only one percentage point less than the ImageNet pretrained network. \textcolor{myblue}{\textsuperscript{\S}}The ImageNet pretrain for VOC uses conv1 through conv5. All fully connected (fc) layers are initialized new.}
\label{table:ablation}
\end{table*}
\begin{table}
\begin{center}
\scriptsize
\begin{tabular}{llllll}
\hline\noalign{\smallskip}
{\bf Method} &	{\bf Class.} &	{\bf Det.} &	{\bf C + D} &	{\bf Seg.} &	{\bf All}\\
\noalign{\smallskip}
\hline
\noalign{\smallskip}
ImageNet Labels \cite{Imagenet09,AlexNet}	& 79.9	& 56.8	& 68.4	& 48.0	& 61.6\\
\noalign{\smallskip}
\hline
\noalign{\smallskip}
Jayaraman \cite{Jayaraman15} & --	& 41.7	& --	& --	& --\\
Li \cite{Li16}	& 56.6	& --	& --	& --	& --\\
Misra \cite{Misra16}	& --	& 42.4	& --	& --	& --\\
Owens \cite{Owens16}\textcolor{myblue}{\textsuperscript{\textdagger}}	& 61.3	& --	& --	& --	& --\\
Larsson	\cite{Larsson17} & 65.9	& --	& --	& 38.4	& --\\
\noalign{\smallskip}
\hline
\noalign{\smallskip}
Agrawal \cite{Agrawal15}\textcolor{myblue}{\textsuperscript{\textdagger}}	& 54.2	& 43.9	& 49.1	& --	& --\\
Gomez \cite{Gomez17}	& 55.7	& 43.0	& 49.4	& --	& --\\
Pathak (Inpainting) \cite{Pathak16}	& 56.5	& 44.5	& 50.5	& 29.7	& 43.6\\
Donahue \cite{Donahue17}\textcolor{myblue}{\textsuperscript{\textdagger}} & 60.1	& 46.9	& 53.5	& 35.2	& 47.4\\
Wang (Video) \cite{Wang15b}\textcolor{myblue}{\textsuperscript{\textdagger}} & 63.1	& 47.4	& 55.3	& --	& --\\
Lee \cite{Lee17}	& 63.8	& 46.9	& 55.4	& --	& --\\
Zhang (Colorizing) \cite{Zhang16}\textcolor{myblue}{\textsuperscript{\textdagger}}	& 65.9	& 46.9	& 56.4	& 35.6	& 49.5\\
Pathak (Move) \cite{Pathak17}	& 61.0	& 52.2	& 56.6	& --	& --\\
Zhang (Split-Brain) \cite{Zhang17}	& 67.1	& 46.7	& 56.9	& 36.0	& 49.9\\
Bojanowski \cite{Bojanowski17}	& 65.3	& 49.4	& 57.4	& --	& --\\
Doersch (Patches)  \cite{Doersch15}\textcolor{myblue}{\textsuperscript{\textdagger}}	& 65.3	& 51.1	& 58.2	& --	& --\\
Noroozi (Counting) \cite{Noroozi17}	& 67.7	& 51.4	& 59.6	& 36.6	& 51.9\\
Noroozi (Puzzle) \cite{Noroozi16b,Noroozi16a}\textcolor{myblue}{\textsuperscript{\textdaggerdbl}}	& 67.6	& \underline{53.2} 		& 60.4	& 37.6	& 52.8\\
Kim \cite{Kim18}\textcolor{myblue}{\textsuperscript{\textbullet}}  & \underline{69.2} & 52.4 & \underline{60.8} & \underline{39.3} & \underline{53.6}\\
\noalign{\smallskip}
\hline
\noalign{\smallskip}
Doersch (Multi-Task)\textcolor{myblue}{\text{*}} \cite{Doersch17}	& --	& \underline{54.9}	& --	& --	& --\\
Wang (Invariance)\textcolor{myblue}{\text{*}} \cite{Wang17}	& --	& 53.1	& --	& --	& --\\
\noalign{\smallskip}
\hline
\noalign{\smallskip}
RWC 180	& 69.0	& 54.9	& 61.9	& 40.4	& 54.8\\
add rotations 90, 270	& 69.5	& 55.5	& 62.5	& {\bf 41.4}	& 55.5\\
add RRM	& {\bf 69.6}	& {\bf 55.8}	& {\bf 62.7}	& 41.2	& {\bf 55.6}\\
\hline
\end{tabular}
\end{center}
\caption{These are classification mAP \cite{Krahenbuhl16}, detection mAP \cite{Girshick15} and segmentation mIU \cite{Long15} test results over PASCAL VOC \cite{Everingham10}. Mean scores are shown for classification + detection \emph{(C + D)} as well as for all three if the segmentation score is available \emph{(All)}. The bottom three results are ours and include all methods except for WV. These are: CB, YJ, TP, EPC, UBT, RA and RWC. RWC (four rotations) gives the best results, but adding in RRM yields only slightly better results. \textcolor{myblue}{\textsuperscript{\textdagger}}To conserve space, we have taken the largest of two scores when network weights have been rescaled \cite{Krahenbuhl16}.  \textcolor{myblue}{\text{*}}Denotes that this is an estimate for the score based on a very recent result with a different network other than AlexNet. The estimate is computed by adding the gain reported in the work to a mutual baseline method that has an AlexNet result and also appears in our table (namely \cite{Doersch15}).  \textcolor{myblue}{\textsuperscript{\textbullet}}Results were published while this paper was under review. \textcolor{myblue}{\textsuperscript{\textdaggerdbl}}Using corrected results from ArXiv paper, not ECCV.}
\label{table:pascal_voc_class}
\end{table}

\begin{table}
\begin{center}
\scriptsize
\begin{tabular}{lllllll}
\hline\noalign{\smallskip}
{\bf Method} &	{\bf C1} &	{\bf C2} &	{\bf C3} &	{\bf C4} &	{\bf C5} &	{\bf Best}\\
\noalign{\smallskip}
\hline
\noalign{\smallskip}
ImageNet \cite{Imagenet09,AlexNet,Zhang16} &	19.3 &	36.3 &	44.2 &	48.3 &	50.5 &	50.5\\
\noalign{\smallskip}
\hline
\noalign{\smallskip}
Random	\cite{Noroozi17} & 11.6 &	17.1 &	16.9 &	16.3 &	14.1 &	17.1\\
\noalign{\smallskip}
\hline
\noalign{\smallskip}
Pathak	(Inpainting) \cite{Pathak16} & 14.1 &	20.7 &	21.0 &	19.8 &	15.5 &	21.0\\
Donahue	\cite{Donahue17} & 17.7 &	24.5 &	31.0 &	29.9 &	28.0 &	31.0\\
Doersch	(Patches) \cite{Doersch15} & 16.2 &	23.3 &	30.2 &	31.7 &	29.6 &	31.7\\
Zhang  (Colorizing) \cite{Zhang16} &	13.1 &	24.8 &	31.0 &	32.6 &	32.6 &	32.6\\
Noroozi  (Puzzle) \cite{Noroozi16b,Noroozi16a} &	\underline{18.2} &	28.8 &	34.0 &	33.9 &	27.1 &	34.0\\
Noroozi  (Counting) \cite{Noroozi17} &	18.0 &	\underline{30.6} &	34.3 &	32.5 &	25.7 &	34.3\\
Kim \cite{Kim18} & 14.5 & 27.2 & 32.8 & 34.3 & \underline{32.9} & 34.3\\
Zhang (Split-Brain) \cite{Zhang17} &	17.7 &	29.3 &	\underline{35.4} &	\underline{35.2} &	32.8 &	\underline{35.4}\\
\noalign{\smallskip}
\hline
\noalign{\smallskip}
RWC 180 &	19.4 &	31.2 &	36.7 &	37.1 &	32.8 &	37.1\\
add rotations 90, 270 &	19.5 &	31.6 &	37.1 &	37.7 &	{\bf 33.7} &	37.7\\
add RRM &	{\bf 19.6} &	{\bf 31.8} &	{\bf 37.6} &	{\bf 37.8} &	33.4 &	{\bf 37.8}\\
\hline
\end{tabular}
\end{center}
\caption{This is the linear test for ImageNet data \cite{Imagenet09}. The network is fine-tuned up to the convolution layer shown. Our results are the bottom three rows. These are the same three self-supervised conditions used in table~\ref{table:pascal_voc_class}. These use all the methods we have presented except for WV.  The maximum score is shown in bold with the previous best result underlined. Rotation with Classification using 90, 180 and 270 degree rotations is generally the best performer.  Here, RRM edges out the other two by a small margin.}
\label{table:linear_imagenet}
\end{table}

\begin{table}
\begin{center}
\scriptsize
\begin{tabular}{lllllll}
\hline\noalign{\smallskip}
{\bf Method} &	{\bf C1} &	{\bf C2} &	{\bf C3} &	{\bf C4} &	{\bf C5} &	{\bf Best}\\
\noalign{\smallskip}
\hline
\noalign{\smallskip}
Places \cite{Places14,Zhang16} &	22.1 &	35.1 &	40.2 &	43.3 &	44.6 &	44.6\\
ImageNet \cite{Imagenet09,AlexNet,Zhang16} & 22.7 &	34.8 &	38.4 &	39.4 &	38.7 &	39.4\\
Random \cite{Noroozi17} & 15.7 &	20.3 &	19.8 &	19.1 &	17.5 &	20.3\\
\noalign{\smallskip}
\hline
\noalign{\smallskip}
Pathak (Inpainting) \cite{Pathak16} 	& 18.2 &	23.2 &	23.4 &	21.9 &	18.4 &	23.4\\
Wang (Video) \cite{Wang15b} &	20.1 &	28.5 &	29.9 &	29.7 &	27.9 &	29.9\\
Zhang (Colorizing) \cite{Zhang16} &	16.0 &	25.7 &	29.6 &	30.3 &	29.7 &	30.3\\
Donahue	\cite{Donahue17} & 22.0 &	28.7 &	31.8 &	31.3 &	29.7 &	31.8\\
Owens	\cite{Owens16}  & 19.9 &	29.3 &	32.1 &	28.8 &	29.8 &	32.1\\
Doersch	(Patches) \cite{Doersch15}  & 19.7 &	26.7 &	31.9 &	32.7 &	30.9 &	32.7\\
Zhang (Split-Brain) \cite{Zhang17} &	21.3 &	30.7 &	34.0 &	34.1 &	\underline{32.5} &	34.1\\
Noroozi (Puzzle) \cite{Noroozi16b,Noroozi16a} & 	23.0 &	31.9 &	35.0 &	34.2 &	29.3 &	35.0\\
Noroozi (Counting) \cite{Noroozi17} &	\underline{23.3} &	\underline{33.9} &	\underline{36.3} &	\underline{34.7} &	29.6 &	\underline{36.3}\\
\noalign{\smallskip}
\hline
\noalign{\smallskip}
RWC 180	 & {\bf 23.7} &	33.9 &	37.1 &	{\bf 37.2} &	34.1 &	{\bf 37.2}\\
add rotations 90, 270	 & 23.5 &	34.0 &	{\bf 37.2} &	{\bf 37.2} &	{\bf 34.9} &	{\bf 37.2}\\
add RRM	 & 23.5 &	{\bf 34.2} &	{\bf 37.2} &	37.0 &	34.4 &	{\bf 37.2}\\
\hline
\end{tabular}
\end{center}
\caption{This is the linear test for CSAIL Places \cite{Places14} data. The network is fine-tuned up to the convolution layer shown. Our results are the bottom three rows. These are the same three self-supervised conditions used in table~\ref{table:pascal_voc_class}. These all use the methods we have presented except for WV.  The maximum score is shown in bold with the previous best underlined. RWC (four rotations) is slightly better, but all three variations obtain the same max score.}
\label{table:linear_places}
\end{table}

\section{Experiments}

We perform a variety of experiments. We show the ablation gain of each tool on our CUB/CCars dataset combination, and also \emph{post hoc} on VOC classification.

\subsection{Self-supervised Training}

We use triple Siamese networks which share weights between branches, and are then concatenated together and run through a few fully connected layers. The input is a set of three 96x96 RGB image patches processed from 110x110 patches, taken from the ImageNet training dataset. Recall that we apply the chroma blur operation offline before we train to remove the expense of repeated Lab conversion and blurring. The output is a softmax classification for the patch arrangement pattern class with 20, 40 or 80 classes. All networks load in a list of shuffled training and testing patches. The list is reshuffled after each epoch. 

We use a slightly different protocol for training the batch normalized networks than for training the non-normalized CaffeNet. The batch normalized networks train with stochastic gradient descent (SGD) for 750k iterations with a batch size of 128 and an initial learning rate of 0.01. A step protocol is used with a size of 300k and gamma 0.1. Momentum is 0.9 and weight decay is 0.0002. For our CaffeNet, we use a Google exponential style training \cite{Inceptionv4}. We train for 1.5 million iterations with a batch size of 128 and initial learning rate of 0.00666 (the fastest rate seemingly stable). We train SGD with a step size of 10k and a gamma of 0.96806. Momentum is 0.9 and weight decay is 0.0002. 

\subsection{Validation and Ablation on CUB Birds and CompCars}

The bulk of our testing and validation was carried out by fine-tuning a self-supervise trained network to the CUB/CCars datasets. Both sets were split \emph{a priori} into training and testing sets by the authors. We use provided bounding boxes from both sets to pre-crop the images. Some further details can be seen in the appendix \ref{compcars_appendix}.

We perform most ablation and validation experiments on our custom batch normalized AlexNet which can be seen in figure ~\ref{fig:alexnet}. The target network is similar to \cite{Doersch15} in that we use the same conv6 and conv6b layers, but we do not try to transfer these layers from the self-supervise trained network. We kept these layers mostly for diversity, so that our batch normalized AlexNet is somewhat different from the very standard CaffeNet/AlexNet we perform benchmark tests on. Again, generalization is important to us. We self-supervise train, then transfer the weights of the five convolution and batch norm layers to the non-Siamese network and initialize new fully connected layers. Both CUB and CCars are trained the same way. The methods for training both had been established \emph{a priori} to avoid over-tuning of hyperparameters. For fine-tuning, we use a polynomial learning policy with an initial learning rate of 0.01 with SGD for 100k iterations with a batch size of 64. Polynomial power is 0.5. Momentum is 0.9 and weight decay is 0.0002. For each condition we wished to test, we trained three times and took the average testing accuracy to reduce minor variation within condition results. 

Ablation results for each method can be seen in Table~\ref{table:ablation}. We show \emph{post hoc} results from PASCAL VOC 2007 classification on the same network and condition to see how well our validation set results map to one of our target data sets. VOC was trained by the standard classification method described in \cite{Krahenbuhl16} and results are in mean average precision (mAP). Finer details on ablation and \underline{more experiments} can be seen in the \emph{appendix} \ref{ablation_appendix}.

\subsection{Standard Transfer Learning Testing Battery}

We demonstrate how the results we have obtained compared with self-supervised methods using a suite of standard benchmark tests. These include classification \cite{Krahenbuhl16}  and detection \cite{Girshick15} on PASCAL VOC 2007, and segmentation \cite{Long15,Pathak16} on PASCAL VOC 2012. They also include the ``linear classifier'' tests on CSAIL places and ImageNet \cite{Zhang16}. We note two possible differences from the standard benchmark methodology here. For detection, we use multiscale training and testing. This is common and used by \cite{Pathak17,Noroozi16a,Noroozi16b,Doersch15,Doersch17}, but not all authors use it. For segmentation, most authors use surgery to map trained fully connected layers to convolution layers six and seven. Our trained network does not have the correct number of weights in the fully connected layers to do this. So we only copy convolution layers one through five and initialize layers six and seven randomly.   

For these tests, we self-supervise train a triple Siamese CaffeNet type AlexNet using the non-batch normalized protocol previously described. Inputs are padded by 5 pixels, but \emph{only} during the self-supervised triple network training. No batch normalization is used at any stage of training. After pooling layer 5, we use the same Siamese structure as our custom batch normalized AlexNet. We leave out batch normalization in these layers, but insert dropout layers after joined\_fc1 and joined\_fc2 with a dropout ratio of 0.5. Convolution weights from layers one through five are transferred to a completely off the shelf CaffeNet.  Training and testing are performed in the standard way defined by the authors of each test (with the two noted differences). Results can be seen in tables~\ref{table:pascal_voc_class}, ~\ref{table:linear_imagenet} and ~\ref{table:linear_places}. Our improvements yield results that out-perform all other methods on all of the standard benchmark tests.


\subsection{Portability to Other Networks}
\begin{table}
\begin{center}
\scriptsize
\begin{tabular}{llll}
\hline\noalign{\smallskip}
{\bf Method} &	{\bf AlexNet BN} &	{\bf ResCeption} &	{\bf Inception 21k}\\
\noalign{\smallskip}
\hline
\noalign{\smallskip}
ImageNet Pretrain &	79.92 &	88.62 &	89.01\\
add Rotations 180 (RWC) &	76.56 &	86.37 &	85.20\\
add rotations 90, 270 (RWC) &	76.82 &	86.52 &	85.81\\
Diff from ImageNet &	3.10 &	2.10 &	3.20\\
\hline
\end{tabular}
\end{center}
\caption{These are the mean results for CUB and CompCars on the different networks.}
\label{table:networks}
\end{table}
We trained on two more networks to demonstrate portability and generalization. The first new network, ResCeption \cite{Mundhenk2016} is a GoogLeNet \cite{GoogLeNet} like network with batch normalization (BN) \cite{BatchNorm} and residual short-cutting \cite{ResNet}. It has 5x5 convolutions in group layer 5 which extend beyond the self-supervised receptive layer. So we self-supervise trained by replacing these with 1x1 surrogate filters that cannot train. Then we put freshly initialized 5x5 convolutions back in place for this layer when we fine-tuned.

We also used a standard inception network with BN. The ImageNet pre-train was performed by \cite{Imagenet21k} on the full set of 21k ImageNet labels. The network required no augmentation. All weights are copied from self-supervised training except for the very top fully connected layer which would be discarded anyway.  Table ~\ref{table:networks} shows the results from these new networks with our BN AlexNet. CompCars results tend to be within about one to two percentage point of ImageNet supervised training. However, CUB runs from three to six.  The results from self-supervision seem enticingly close to ImageNet supervised, but are not yet there. 


\section*{Acknowledgments}
The authors would like to thank several people who contributed ideas and conversation for this work: Carmen Carrano, Albert Chu, Alexei Efros, Gerald Friedland, Will Grathwohl, Brenda Ng, Doug Poland, Richard Zhang, Miles the Cat and the CVPR reviewers. This work was performed under the auspices of the U.S. Department of Energy by Lawrence Livermore National Laboratory under Contract DE-AC52-07NA27344 and was supported by the LLNL-LDRD Program under Project No. 17-SI-003. Support was also provided by the LLNL DSSI Summer Institute. 

{\small
\bibliographystyle{ieee}
\bibliography{egbib}
}


\clearpage
\appendix
\section{Appendix: CompCars Dataset Augmentation} \label{compcars_appendix}

\begin{figure*}
\centering
\includegraphics[height=22 cm]{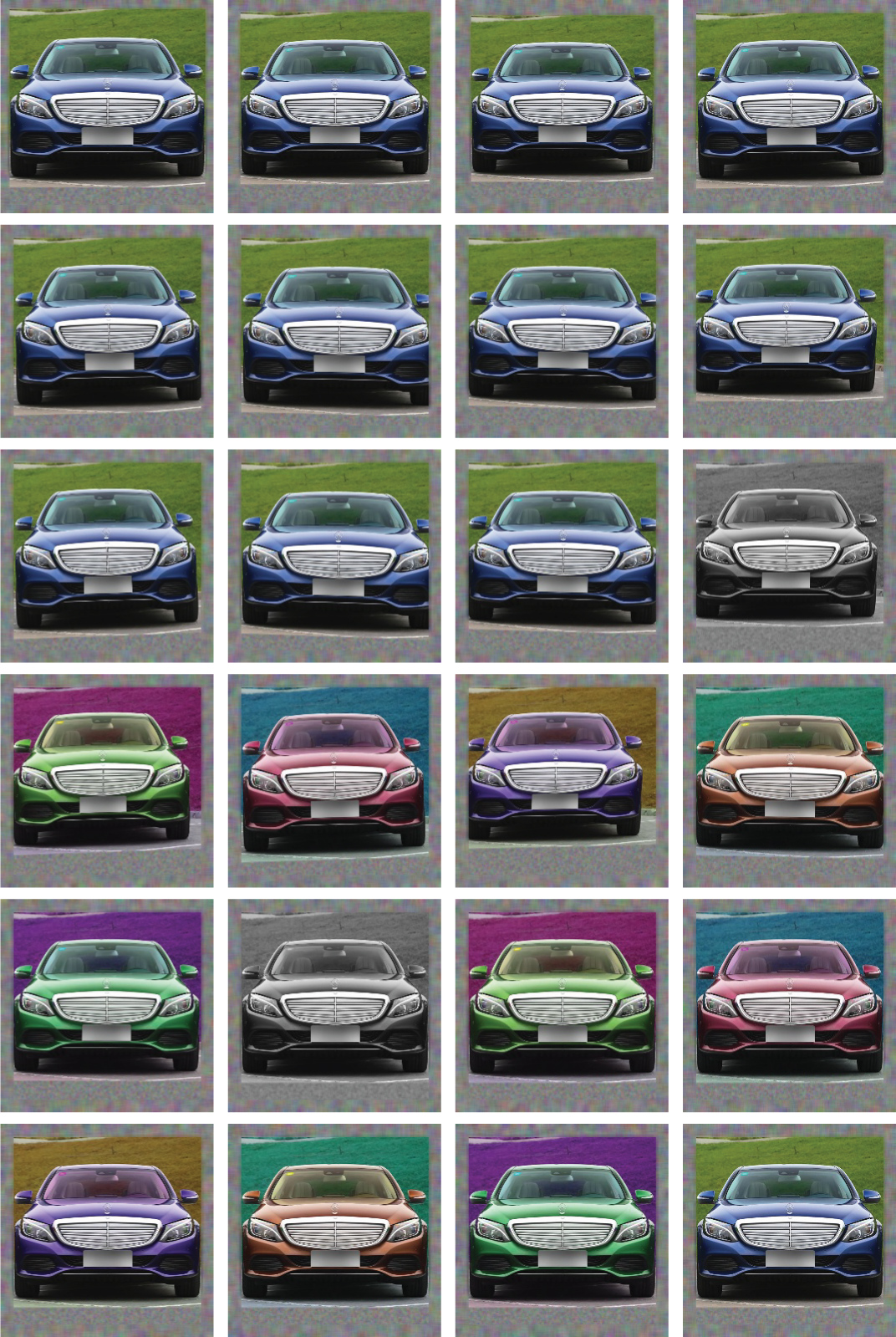}
\caption{These show all 24 augmentations for a single image in CompCars. These variations are applied to all training images. }
\label{fig:compcars}
\end{figure*}

We further augment the CompCars dataset by creating rotating hue and minor perspective jitter. In prior unpublished experiments, these changes seemed to improve accuracy. We rotate the hue by simply swapping color channels. We do this because we hypothesized that car models have varying color, but they seem to have color styles. For instance, family sedans seem to have conservative low saturated colors while sports cars tend to have hot intense and highly saturated colors. We obtain perspective jitter by randomly perturbing three Euler angles by +/- 0.00286 degrees, then we create a perspective transformation matrix from it. We create 24 augmentations per image, from which 14 have random perspective jitter and 12 have hue rotation. Six of the hue rotated images also have perspective jitter. Four images are just a repeat of the un-augmented original. An example of these alterations can be seen in figure ~\ref{fig:compcars}. We create these permutations before training since perspective transformation is modestly expensive, but still can be a bottleneck.
\begin{figure*}
\centering
\includegraphics[height=9.4 cm]{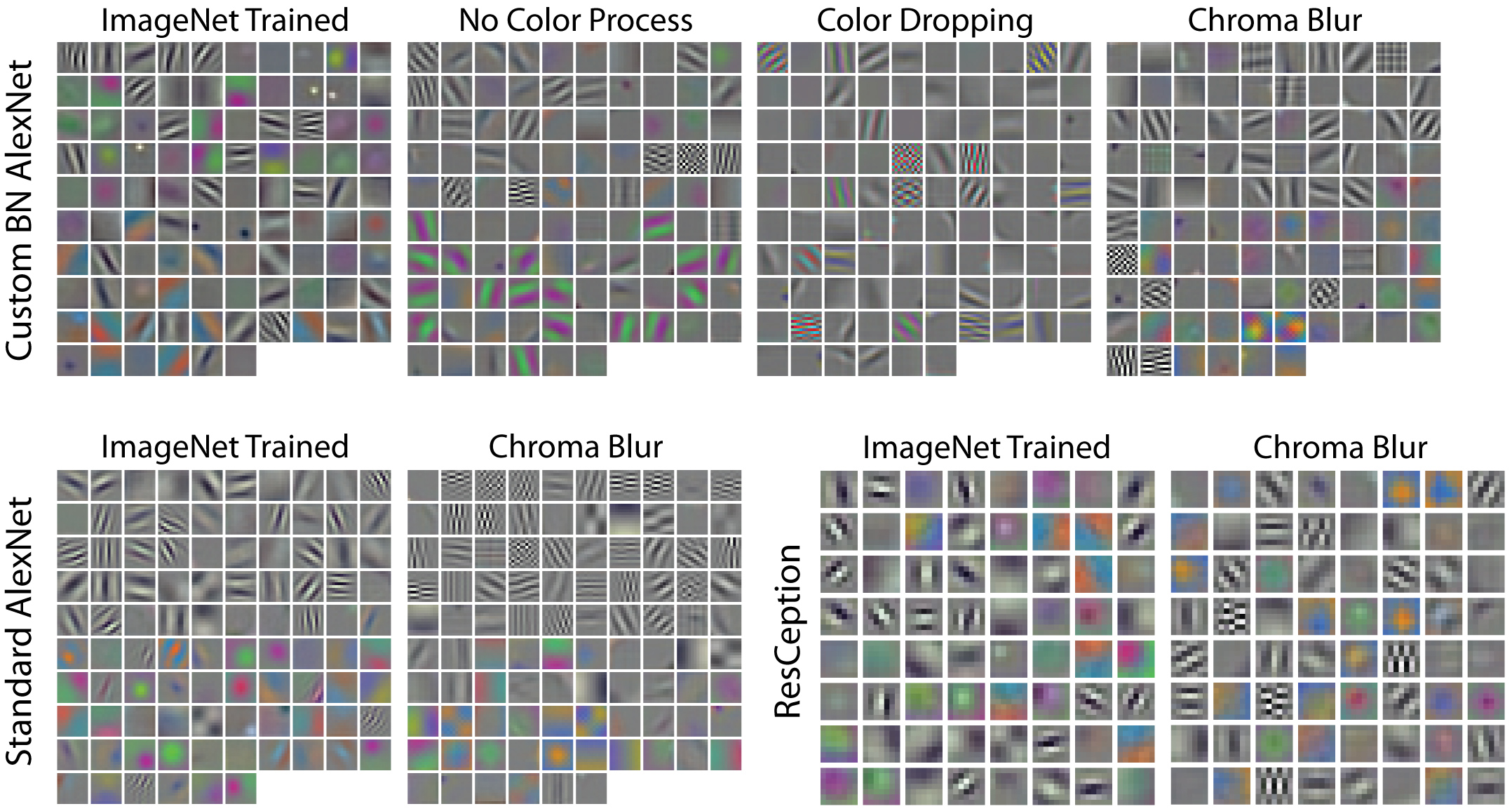}
\caption{These are the first layer filters from several networks which have either been supervise trained on ImageNet or self-supervise trained. All self-supervised networks are using the full set of methods (but not RRM or WV). Even though rotation with classification may mitigate chromatic aberration, the network still forms filters sensitive to it. Thus, we believe it is only a partial solution. The chroma blur networks are all free of chromatic aberration effects and show formation of healthy color filters (especially when compared to color dropping). As an observation, we can zoom to an effective size of 171x171 during self-supervised training; as a result, we see the presence of finer wavelets compared with ImageNet training.}
\label{fig:chroma_blur_layer}
\end{figure*}
\begin{figure}
\centering
\includegraphics[height=5.5 cm]{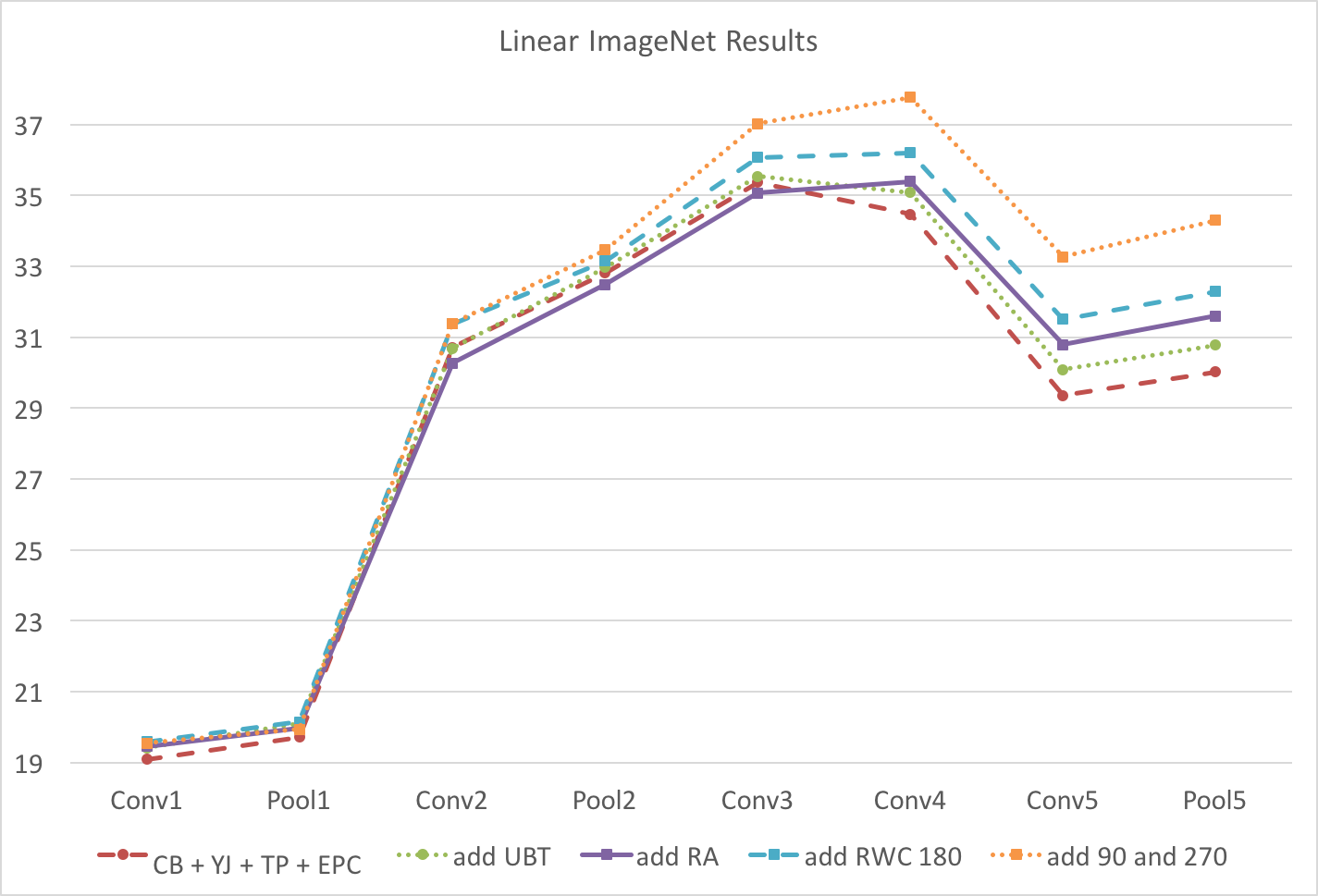}
\caption{These are linear results for CaffeNet/AlextNet on ImageNet. We can see that when we add Random Aperture (RA), the results seem to switch over to improving layers four and five at the expense of layers one, two and three. Note that we did not use padding during self-supervised training for this experiment.}
\label{fig:linear_imagenet}
\end{figure}
\section{Appendix: Further Ablation Details} \label{ablation_appendix}

\subsection{Yoked Jitter Gets Better with Extra Patches}

As mentioned, one of the goals of using hybrid patches was to reduce the ability of the network to learn trivial pattern completion between adjacent patches. As a test, we tried the usage of extra patch configurations (EPC) with yoked jitter and with random jitter. The results can be seen in Table  ~\ref{table:yoked_jitter}.  By getting a larger gain for yoked jitter, there is some evidence that EPC may have the effect of reducing trivial pattern completions.

\subsection{Do Rotations Need Classification?}
We tested to see if just rotating a patch without classification for that rotation was sufficient to improve performance. Table ~\ref{table:rotation_w_classification} shows that if we just rotate the patch, there is almost no difference than without rotation. The classification component might help to sharpen the features of objects by forcing the network to recognize the rotated objects uniquely. Also, as we have discussed, classification may help to mitigate chromatic aberration.
\begin{table}
\begin{center}
\scriptsize
\begin{tabular}{lllll}
\hline\noalign{\smallskip}
{\bf Method} &	{\bf CUB} &	{\bf CCars} &	{\bf Mean } &	{\bf Improvement}\\
\noalign{\smallskip}
\hline
\noalign{\smallskip}
CB &		64.29 &	80.80 &	72.55 & --\\	 
CB + YJ &	65.17 &	80.95 &	73.06 &	0.51\\
\noalign{\smallskip}
\hline
\noalign{\smallskip}
CB + TP + EPC &	65.21 &	80.17 &	72.69 & --\\	 
CB + TP + EPC + YJ &	67.07 &	80.50 &	73.79 &	1.10\\
\hline
\end{tabular}
\end{center}
\caption{We get a general improvement from using a yoked jitter over a random jitter. When we then include the extra patch configurations, the improvement grows. The hybrid patches intrinsically may prevent low-level trivial boundary completion since they have mixed scales.}
\label{table:yoked_jitter}
\end{table}
\begin{table}
\begin{center}
\scriptsize
\begin{tabular}{llll}
\hline\noalign{\smallskip}
{\bf Method} &	{\bf CUB} &	{\bf CCars} &	{\bf Mean }\\
\noalign{\smallskip}
\hline
\noalign{\smallskip}
CB + YJ + TP + EPC + UBT + RA & 	68.01 &	82.07 &	75.04\\
... + RWC 180 without classification &	68.26 &	81.82 &	75.04\\
... + RWC 180 with classification &	68.89 &	84.23 &	76.56\\
\hline
\end{tabular}
\end{center}
\caption{By just rotating the patches but not classifying them, we obtain almost no gain. It appears critical that rotations should have their own class. Note we are using the full toolset except for RRM or WV.}
\label{table:rotation_w_classification}
\end{table}

\subsection{How much does Chroma Blurring Help?}

As figure ~\ref{fig:chroma_blur_layer} shows, chroma burring definitely seems to remove any chromatic aberration effects while preserving at least some color feature processing. Looking at Table ~\ref{table:chroma_blur}, we do get a moderate boost in classification by using chroma blurring compared with no color processing. However, improvement in classification from chroma blurring subsides when all tools are used. We believe that rotation with classification is probably responsible for this.

\begin{table}
\begin{center}
\scriptsize
\begin{tabular}{lllll}
\hline\noalign{\smallskip}
{\bf Method} &	{\bf CUB} &	{\bf CCars} &	{\bf Mean } &	{\bf Impr.}\\
\noalign{\smallskip}
\hline
\noalign{\smallskip}
YJ &	65.04 &	80.21 &	72.62 & --\\	 
... + CD &	62.36 &	79.70 &	71.03 & --\\	 
... + CB &	65.17 &	80.95 &	73.06 &	0.44\\
\noalign{\smallskip}
\hline
\noalign{\smallskip}
YJ + TP + EPC + UBT + RA + RWC &	68.23 &	83.70 &	75.96 & --\\	 
... + CD &	65.73 &	82.94 &	74.33 	& --\\ 
... + CB &	68.42 &	83.58 &	76.00 &	0.04\\
\hline
\end{tabular}
\end{center}
\caption{Here we compare color dropping (CD) and chroma blurring (CB) to no color processing. CUB birds is a pathological case for color dropping since it is very dependent on color patterns for classification. However, and somewhat perplexingly, color dropping does not appear to help CompCars either.  Chroma blurring ceases to help once we add in the full set of tools (not including RRM or WV). We suspect this is because rotation with classification at least partially mitigates the effects of chromatic aberration.}
\label{table:chroma_blur}
\end{table}

\begin{table}
\begin{center}
\scriptsize
\begin{tabular}{lllll}
\hline\noalign{\smallskip}
{\bf Method} &	{\bf CUB} &	{\bf CCars} &	{\bf Mean } &	{\bf Impr.}\\
\noalign{\smallskip}
\hline
\noalign{\smallskip}
CB + YJ + TP & 	65.19 &	81.54 &	73.36 & --\\	 
... + EPC (2x2 Patches) &	66.80 &	81.80 &	74.30 &	0.93\\
... + EPC (Hybrid Patches) &	67.32 &	81.69 &	74.50 &	1.14\\
... + EPC (2x2 \emph{and} Hybrid Patches) &	67.07 &	80.50 &	73.79 &	0.43\\
\noalign{\smallskip}
\hline
\noalign{\smallskip}
CB + YJ + TP + UBT + RA + RWC &	68.63 &	82.87 &	75.75 & --\\	 
... + EPC (2x2 Patches) &	68.29 &	83.67 &	75.98 &	0.23\\
... + EPC (Hybrid Patches) &	67.09 &	82.58 &	74.83 &	\textcolor{myred}{-0.92}\\
... + EPC (2x2 \emph{and} Hybrid Patches) &	68.89 &	84.23 &	76.56 &	0.81\\
\hline
\end{tabular}
\end{center}
\caption{Here we show the effect of the two types of extra patch configurations on their own. Improvement is not straight forward from the addition of the two types. Using both is always better than using just the 3x3 patches. However, when using the full tool set (not including RRM or WV), the hybrid patches by themselves are actually worse. Our hypothesis is that the 2x2 patches help with rotation classification (RWC) since they always include the top and bottom of the image. These are good locations for cues an image is upside-down (sky v. ground). Without that help, the hybrid patches somehow inhibit performance. It is not entirely clear why.}
\label{table:types_of_patches}
\end{table}
\subsection{The Benefit of Adding Different Kinds of Patches}

Extra patch configurations definitely seem to help, but their interaction with each other and the other tools is not deterministic. Table ~\ref{table:types_of_patches} parses out the contribution of the two types of new configurations we use.

\subsection{Two v. Three Apertures}
We chose to only apply the patch aperture to two patches and not all three in a set. The idea was to inhibit the highest levels of the network and instead focus learning on mid-levels. If we applied the aperture to all three patches, we reasoned that we would \emph{always} inhibit the higher levels when we only want to inhibit them \emph{some of the time}. As table ~\ref{table:two_three_aperture} shows, two apertures are definitely better than three. 

\begin{table}
\begin{center}
\small
\begin{tabular}{llll}
\hline\noalign{\smallskip}
{\bf Method} &	{\bf CUB} &	{\bf CCars} &	{\bf Mean }\\
\noalign{\smallskip}
\hline
\noalign{\smallskip}
With three apertures &	67.76 &	83.22 &	75.49\\
With two apertures &	69.04 &	83.46 &	76.25\\
\hline
\end{tabular}
\end{center}
\caption{If we aperture all three patches in a set, we see a noticeable drop particularly in CUB Birds. We did not test the aperture of one patch.}
\label{table:two_three_aperture}
\end{table}

We ran some further testing to see if applying the aperture has the effect of improving middle layers. Figure ~\ref{fig:linear_imagenet} shows that it has a boost on layers four and five which are middle layers in AlexNet. Interestingly, it has somewhat degrading effects on layers one and two. This is a kind of behavior we would expect if mid-layers are being biased for. 

\begin{figure*}
\centering
\includegraphics[height=6.2 cm]{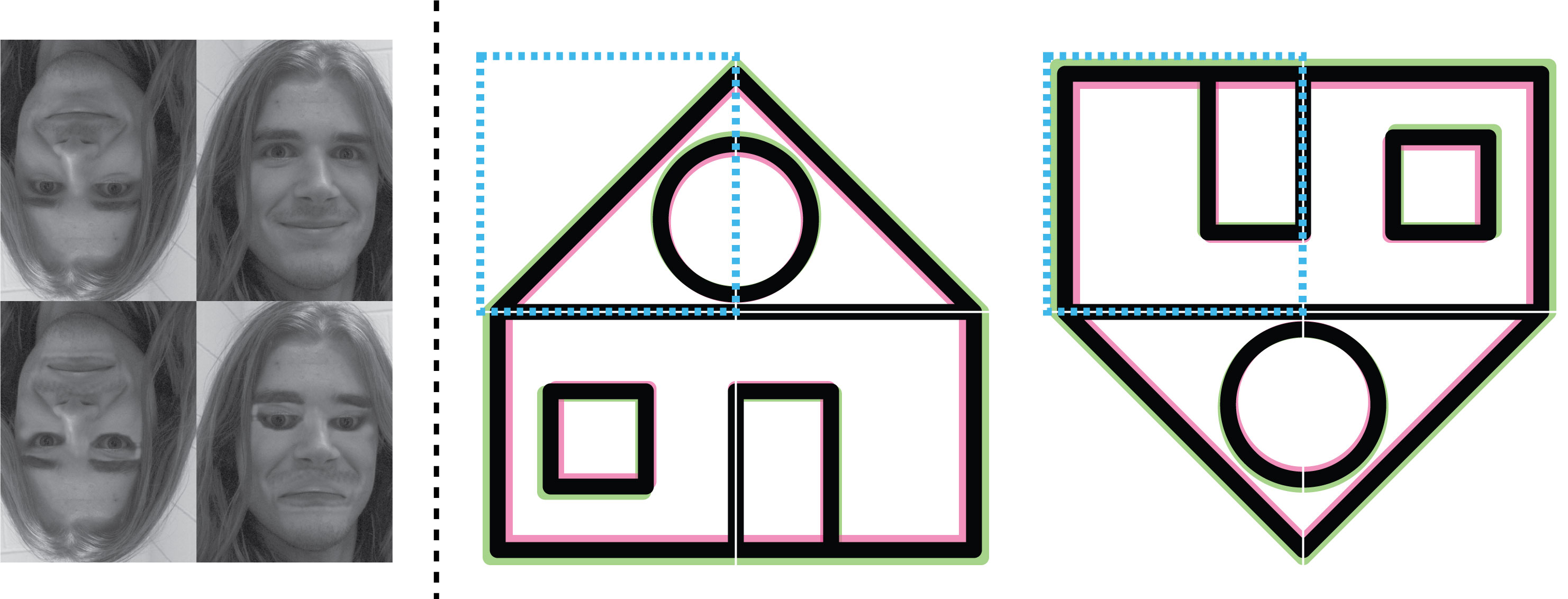}
\caption{This is figure~\ref{fig:rwc} enlarged. On the left is an example of the famous Thatcher illusion \cite{Thompson80,MTeffect06}. It demonstrates conditional sensitivity to upside-down features in an image against the background. We used this mostly as inspiration. On the left house image \cite{ChromaHouse}, the network can tell that the blue bordered area comes from the upper left corner based on chromatic aberration alone. However, on the right image, rotation with classification makes it tell us if the patch is inverted and comes from the lower right corner. If it uses chromatic aberration as the only cue, it would be wrong 50\% of the time.}
\label{fig:rwc_big}
\end{figure*}
\begin{figure*}
\centering
\includegraphics[height=8.6 cm]{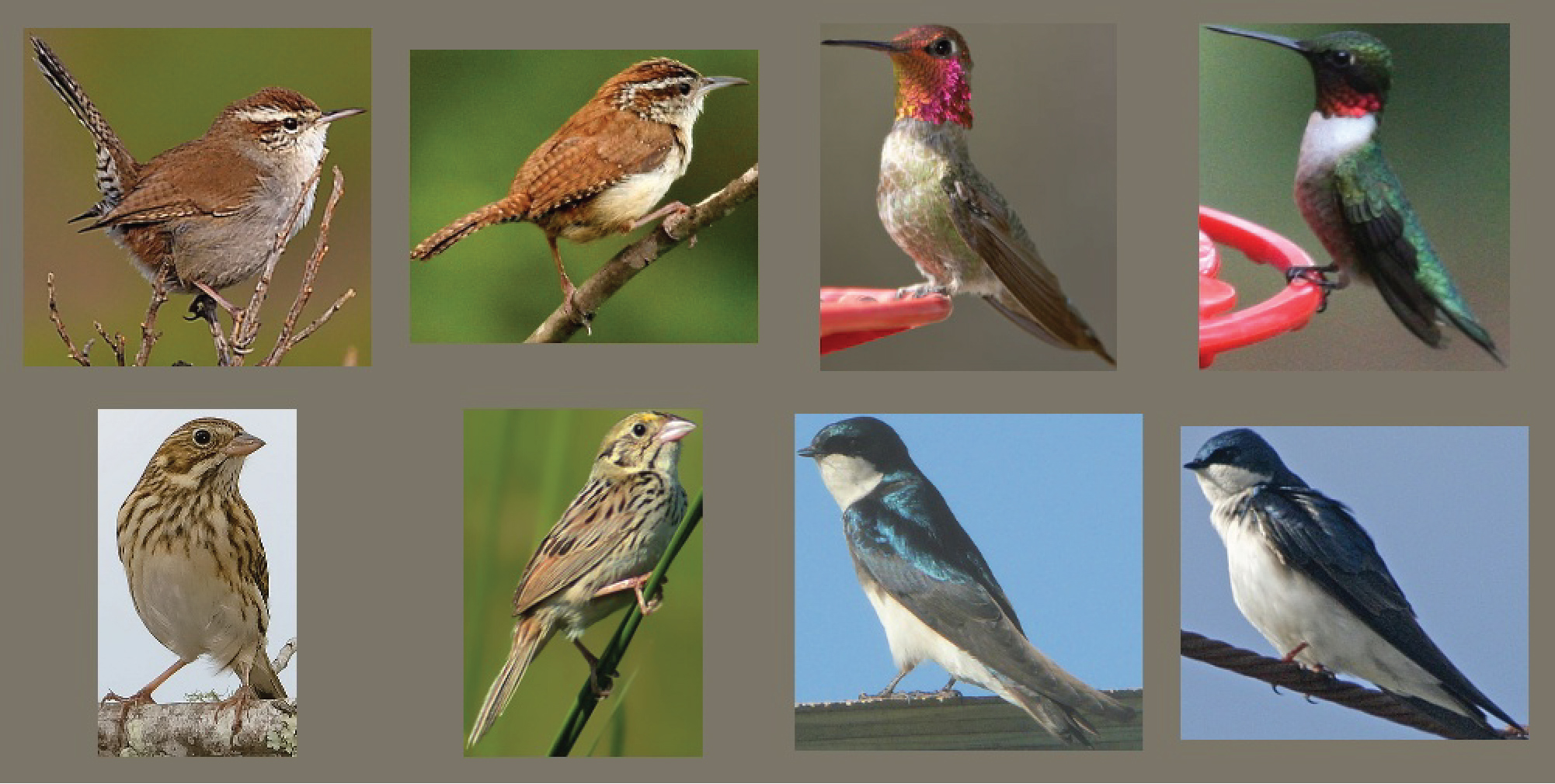}
\caption{These are examples of birds in the CUB birds dataset. Each one is a different species. They are a Bewick Wren, Carolina Wren, Anna Hummingbird, Ruby Throated Hummingbird, Vesper Sparrow, Henslow Sparrow, Tree Swallow and a Bank Swallow.}
\label{fig:birds}
\end{figure*}
\subsection{RRM is a Tiny Bit Better}

Randomization of rescaling methods (RRM) yielded slightly better results on the ImageNet linear and VOC tests. It is roughly even on the CSAIL places linear test. Earlier experiments showed a stronger pattern of gain, but it is now somewhat unclear how much it helps. However, it doesn't seem to hurt. 

\section{Appendix: Document Revision History} \label{revision_appendix}
{\bf Version 1} : \emph{Posted 17 Nov 2017}.

{\bf Version 2} : \emph{Posted 2 Jan 2018}. We revised segmentation results by using the standard method from \cite{Long15}. Our prior results used a method which deviated slightly from it. We also fixed a flaw in the self-supervised Inception model. The connection between the triple network trunk and the concatenated model was supposed to be the same as the design for ResCeption. This improved the results slightly. 

{\bf Version 3} : This adds changes to the text based on reviewers comments at CVPR. This revision is also the one which will appear in the CVPR proceedings. The URL on the front page was updated to reflect the new supported domain for ``Green Data Oasis'' LLNL web hosting. We added results of \cite{Kim18} that came out after our draft version 2. Also, we added padding to the CaffeNet model but \emph{only} to the self-supervised pre-train. The transfer learning CaffeNet (e.g. for VOC testing) stays vanilla. Padding should have been there before, but we missed it. This noticeably improves VOC detection and segmentation results, but has little effect on classification. However, this only applies to CaffeNet since by default the other newer networks (e.g. Inception) all use padding. For the curious, ArXiv version 2 of this document (\url{https://arxiv.org/abs/1711.06379v2}) has the pre-padding results.  

\end{document}